\documentclass{svproc}
\usepackage{url}
\usepackage{amsmath}
\usepackage{graphicx}
\usepackage{float}
\usepackage{booktabs}
\usepackage{multirow}
\usepackage{array}
\usepackage{tensor}
\usepackage{hyperref}

\begin{document}
\mainmatter        
\title{Evaluation of facial landmark localization performance in a surgical setting}
\titlerunning{Facial landmark localization in surgical setting}  
\author{Ines Frajtag\inst{1} \and Marko Švaco\inst{1}\and Filip Šuligoj\inst{1} }
\authorrunning{Ines Frajtag et al.} 
\tocauthor{Marko Švaco, Filip Šuligoj}
\institute{Faculty of Mechanical Engineering and Naval Architecture, University of Zagreb, Croatia\\
\email{ines.frajtag@fsb.unizg.hr}\\ 
}

\maketitle             

\begin{abstract}
The use of robotics, computer vision, and their applications is becoming increasingly widespread in various fields, including medicine. Many face detection algorithms have found applications in neurosurgery, ophthalmology, and plastic surgery. A common challenge in using these algorithms is variable lighting conditions and the flexibility of detection positions to identify and precisely localize patients. The proposed experiment tests the MediaPipe algorithm for detecting facial landmarks in a controlled setting, using a robotic arm that automatically adjusts positions while the surgical light and the phantom remain in a fixed position. The results of this study demonstrate that the improved accuracy of facial landmark detection under surgical lighting significantly enhances the detection performance at larger yaw and pitch angles. The increase in standard deviation/dispersion occurs due to imprecise detection of selected facial landmarks. This analysis allows for a discussion on the potential integration of the MediaPipe algorithm into medical procedures. 

\end{abstract}

\section{Introduction}
Medical practice has advanced significantly with the development of technology and automation, providing new solutions to improve the precision, efficiency, and safety of medical procedures. Traditional surgical methods often rely on the experience of surgeons \cite{han} and visual techniques, which can be prone to human error. Modern intelligent solutions such as robotic systems \cite{hai}, computer vision algorithms, and precise planning \cite{jerbic} are introduced in a surgical setting. 

One of the key computer vision technologies is facial landmark detection, which enables the identification of specific anatomical features such as eyes, eyebrows, nose, and mouth. In medical applications, particularly plastic surgery \cite{ibrahim}, neurosurgery \cite{jerbic} or ophthalmology \cite{simsek}, tracking landmark positions is crucial for determining precise surgical sites. Most face detection algorithms, such as OpenPose \cite{balt}, MediaPipe \cite{medi}, Dlib library \cite{king}, and FaceNet \cite{schroff}, are available as open-source tools and are widely used for tasks such as emotion recognition \cite{jakhete}, surveillance, and identification. However, the application of these algorithms in operating rooms remains underexplored. The implementation of face recognition algorithms and facial landmark detection in operating rooms facilitates a seamless process to identify the target area for surgery in the patient. Not only does this expedite the surgical procedure, it enhances its safety by allowing continuous tracking of the face and its features throughout the entire surgical process. A specific research question addressed in this study is the following: How do lighting conditions such as intensity and angle influence the accuracy and repeatability of facial landmark detection in operating rooms?

MediaPipe is an open source framework based on the BlazeFace method \cite{zhu}, and the aim of this study is to evaluate its algorithm to detect facial landmarks under realistic operating room conditions. The experiment involves testing the algorithm in controlled medical conditions using a robotic arm with an attached 3D camera and an anthropomorphic phantom, all set within an operating room. Furthermore, this study seeks to investigate whether MediaPipe can be used to estimate the pose of the patient's head during the planning and execution of surgical procedures. The objective is to assess the consistency of the algorithm in detecting the corresponding facial landmarks at varying angles of capture. The results of this research will contribute to a better understanding of the limitations and reliability of the algorithm in medical scenarios, potentially advancing its application in operating rooms.

\section{Related research}
Lighting-related issues present a significant challenge in the field of face recognition. Conditions such as weak, strong or uneven lighting make it difficult for algorithms to detect geometric features of the face, potentially leading to detection failures or reduced repeatability. In the study \cite{kumar}, the authors combined multiple face detection methods to improve results under varying lighting, skin color, and background conditions. To address lighting issues, they suggest that optimization algorithms are less sensitive to external factors, with neural networks trained on diverse facial illumination datasets that enable better face detection \cite{bao}. Another approach is the multi-exposure concept. In \cite{liang}, the authors showed that their Recurrent Exposure Generation (REG) module could be effectively combined with various face detection algorithms without needing multiple images in low or normal lighting. In \cite{nuaimi}, MediaPipe’s facial landmark recognition and head pose prediction were investigated, showing that it reliably estimates head pose angles using deep neural network methods. However, lighting conditions and head pose angles were identified as key challenges that affect the accuracy of facial landmark detection. Due to the conclusion in \cite{nuaimi}, our study focuses on the impact of lighting and facial landmark detection under realistic clinical conditions. In \cite{hammadi}, a comparison of MediaPipe, OpenPose 2.0, and 3DDFA-V2 was performed using a motion capture system as the gold standard. Based on the findings, 3DDFA-V2 was recommended as the most suitable for clinical applications where the patient moves in various directions. The primary distinction between \cite{hammadi} and our study is the integration of a robotic arm with a 3D camera and a fixed phantom, mimicking a neurosurgical setup. There are differences in lighting (our study uses surgical light) and in setting limits on detection angles, changes in 7 specific facial landmarks are monitored. The position of the predicted facial landmarks was observed using 2D and 3D coordinates, in order to demonstrate their dispersion and the repeatability of localization. In their study \cite{ali}, Abdullah and Ali compared several deep learning-based face detectors. Among MTCNN, Dlib library and MediaPipe, the results showed that MediaPipe achieved the highest accuracy at 99.3\% \cite{ali}. The criteria used to determine accuracy included the type and number of errors in detection, precision, and execution time. MediaPipe can determine 468 interdependent points, even the smallest facial movement, resulting in a change in the position of these points. Due to its sensitivity to shifts and its ease of integration into various systems, it was chosen for this research. Furthermore, the algorithm is frequently applied in surveillance systems and other areas of medicine (rehabilitation \cite{dill}, diagnostics\cite{galanakis}), where it serves for research purposes. The algorithm is used in the interdisciplinary field of medicine and engineering \cite{shan}, where blink counts during driving are compared with medical monitoring methods (EOG and EEG). In the study \cite{shan}, MediaPipe achieved an accuracy of 99.87\% for blink detection and recorded twice as few false blinks compared to the Dlib library.
\section{Materials and Methods}
This chapter outlines the application of the MediaPipe algorithm for facial landmark detection. The primary focus is on assessing the reproducibility of landmark identification and analyzing the performance of the algorithm in relation to position and light variations. These evaluations aim to determine the robustness of the algorithm and its potential suitability for clinical environments.

\begin{figure}[H]
    \centering
    \includegraphics[width=1\linewidth]{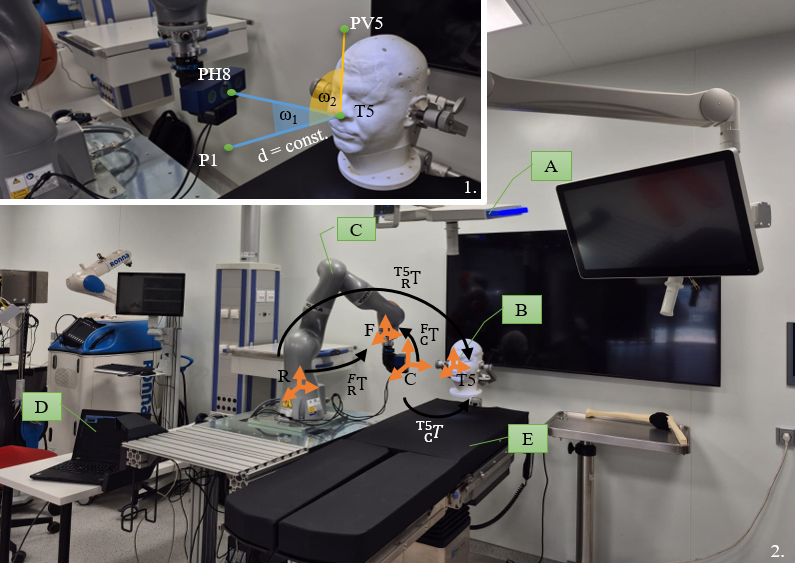}
    \caption{1. General laboratory setup including the surgical light (A), phantom (B), robot with attached 3D camera (C), PC running MediaPipe (D) and operating table (E). Major reference frames are: the robot base (R), robot flange (F), camera (C) and image frames (T5). Homogeneous transformation matrix between any two coordinate frames (y and x) is denoted as \(_Y^{X}T\) \cite{suli}. To achieve the rotated positions of the robot arm for a specific angle we use a rotation matrix \(R_x\) for yaw and \(R_y\) for pitch. 2. The positions from P1 to PH8 are generated using equation \ref{eq:1} for values \(\omega_1=\{0^\circ, 10^\circ, 20^\circ, 30^\circ, 40^\circ, 50^\circ, 60^\circ, 70^\circ\}\). In the same way positions from P1 to PV5 are generated using equation \ref{eq:2} for values \(\omega_2 = \{0^\circ, 10^\circ, 20^\circ, 30^\circ, 40^\circ\}.\)}
    \label{fig:prva}
\end{figure}

The experiment was conducted in the Laboratory for Medical Robotics (Faculty of Mechanical Engineering and Naval Architecture, University of Zagreb) which is designed as a mock-up of an operating room. The equipment of \autoref{fig:prva} (part 2) was used to perform the experiment. The phantom was securely fixed on the DORO QR3 Multi-Purpose Skull Clamp attached to the Hillrom PST 300 operating table. The phantom geometry was derived from reconstruction of CT images of the patient and was fabricated from PLA material using a 3D printer. The phantom was developed using patient data, with the patient's informed consent, and with the approval of the ethics committee for the use of CT images in research. Above the phantom, there is a Trumpf TruLight 5000 surgical light, while the KUKA LBR iiwa 14 R820 robot equipped with a 3D camera Ensenso N35-606-16-IR performs movements to predefined positions. The calculation of positions, command execution, and initiation of the MediaPipe algorithm are conducted on a research workstation running Ubuntu 22.04, featuring a 64-bit Intel Core i5 processor (2.80 GHz). Initial validation and camera positioning are determined based on the coordinates of the phantom's facial landmark for the nose tip (marked T5 in \autoref{fig:druga} (part 2)). The facial landmark coordinates on a 2D image were identified using the MediaPipe FaceMesh algorithm on 10 repeated captures. Data from the disparity maps captured by the 3D camera, combined with the 2D point coordinates, are used to calculate the 3D coordinates. The camera is placed at the distance \textit{d} = 270 mm from the facial landmark T5, this distance remaining constant as the robot and the camera are moved to different positions.

\noindent  The robot positions PH and PV were calculated based on head rotation angles ($\omega_1$ for yaw and $\omega_2$ for pitch) using the eigen C++ library to compute the transformation matrix and using the corresponding equation \autoref{eq:1} or \autoref{eq:2}.
\begin{equation}
    \tensor*[_R^{T5}]{T}{} = _R^{F}T \cdot \left(_C^{F}T \right)^{-1} \cdot_C^{T5}T \cdot R_y(\omega_1) \cdot \left(_C^{T5}T \right)^{-1} \cdot _C^{F}T
    \label{eq:1}
\end{equation}
\begin{equation}
\tensor*[_R^{T5}]{T}{} = _R^{F}T \cdot \left(_C^{F}T \right)^{-1} \cdot_C^{T5}T \cdot R_x(\omega_2) \cdot \left(_C^{T5}T \right)^{-1} \cdot _C^{F}T
\label{eq:2}
\end{equation}
While conducting the experiment, the surgical light was set to maximum intensity (160 000 lux) with a color temperature of 4500 K and the angle between the surgical light and the phantom is 5°. 
Focusing on facial analysis, the MediaPipe FaceMesh model employs a geometric approach to predict 468 facial landmarks, of which seven key points are tracked in the study.

Data collection proceeds as follows: the robot moves the camera to the desired position, waits 5 seconds, captures two images, and processes them using the MediaPipe algorithm. The output includes the rotation angle, the detected facial landmarks, the 2D coordinates of the landmarks, and their respective 3D coordinates. 

\section{Results}
Facial landmark detection performance was evaluated under two different lighting conditions: surgical light and an integrated infrared front light on a 3D camera. A robotic arm was used to position the camera at predefined angles, automating the image capture process.

\begin{figure}[H]
    \centering
    \includegraphics[width=1\linewidth]{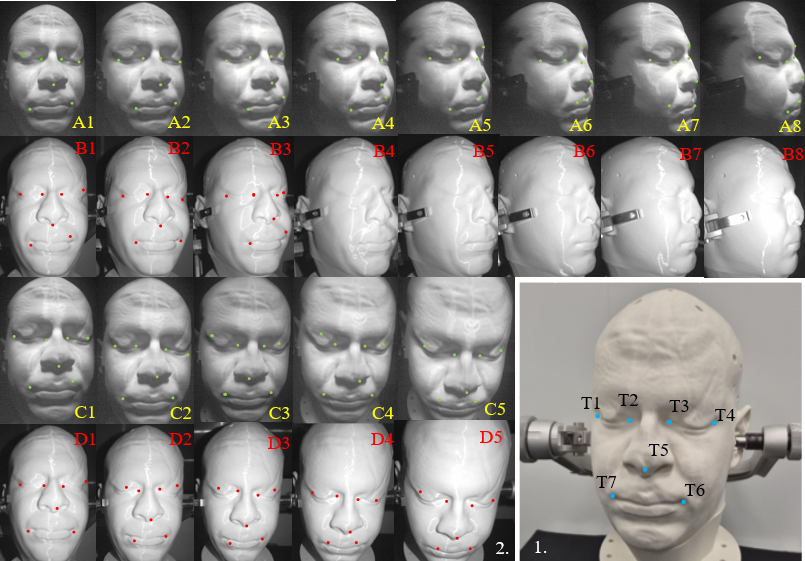}
    \caption{1. The seven specific points that are observed during testing are: T1 (outer corner of the right eye), T2 (inner corner of the right eye), T3 (inner corner of the left eye), T4 (outer corner of the left eye), T5 (nose tip), T6 (left mouth corner), and T7 (right mouth corner). 2. The head positions, rotated by an angle \(\omega_1\) from P1 to PH8, and the detection or non-detection of key points are shown in the images labeled A1-A8 when only the surgical light was used. The same positions are shown and labeled B1-B8 when the integrated front light from the camera was used. Similarly, head positions rotated by an angle \(\omega_2\) from P1 to PV5 are presented, with C1-C5 indicating images where only the surgical light was used, and D1-D5 indicating images where the front light from the camera was used.}
    \label{fig:druga}
\end{figure}

\begin{figure} [H]
    \centering
    \includegraphics[width=1\linewidth]{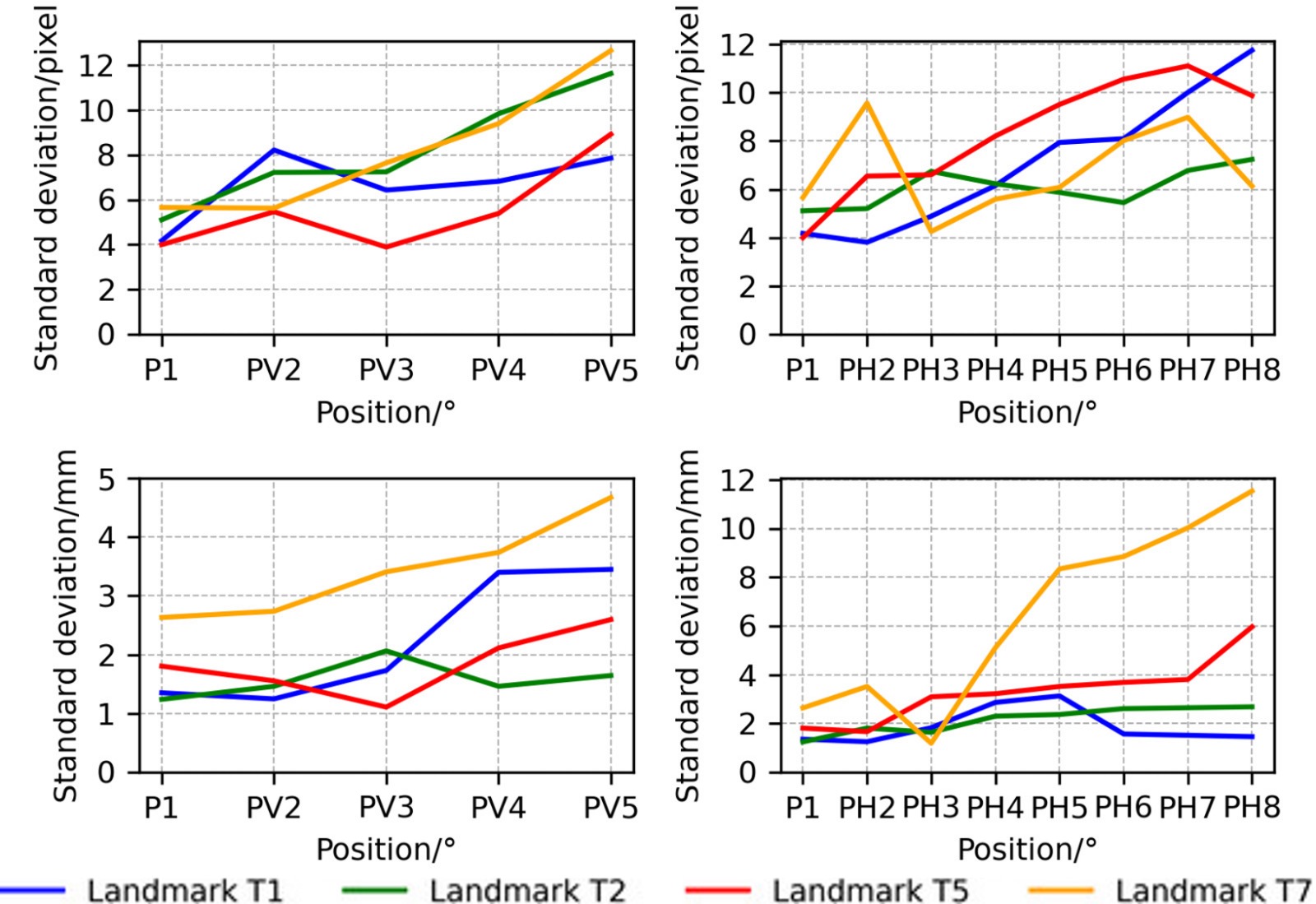}
    \caption{The graphs show the relationship between standard deviation and positions. The upper graphs display the results for the image coordinate in pixels, while the lower graphs show results for 3D coordinates. On the left side are the graphs for the yaw direction and on the right side are the graphs for the pitch direction.}
    \label{fig:grafovi}
\end{figure}

To identify detection limits in scenarios where surgical light is used (the front light of the camera is off), the robotic arm performed rotations in 10° increments. After detection failed beyond a rotation of 80° in yaw and 40° in pitch, testing was carried out with 1° increments. These tests determined that the algorithm stopped detection at 73° and 40°. The repeatability of facial landmark detection at 73° was 40\% out of 10 attempts. The camera was moved to 7 positions in yaw and 5 positions in pitch, at predefined angles ($\omega_1$ for yaw and $\omega_2$ for pitch), see \autoref{fig:prva} (part 1). For each position, 10 tests were conducted. The results demonstrated that detection was successful for all positions, including yaw 70° and pitch 40°. Using the front light on the camera, which uses an infrared projector to illuminate the scene, the maximum angle for face detection is significantly reduced. The test revealed that the algorithm detected the face at 30° in only 30\% of the cases. However, at 24°, the facial landmark predictions were 100\% successful. Since MediaPipe predicts points after detecting a face, not all landmarks were accurately identified in positions where they are supposed to be. For this reason, the coordinate changes were monitored for 4 landmarks: T1, T2, T5 and T7, which were successfully detected in all positions. In the first phase, the maximum achievable rotation angle for face detection was tested using variable surgical lighting and the integrated front light of a 3D camera. The results are shown in \autoref{fig:druga} (part 2). In the second phase, changes in the coordinates value of the landmarks that were consistently and accurately detected were tracked. Based on the results, under constant lighting conditions, an increase in the rotation angle leads to a greater dispersion of the coordinates of a specific landmark, as indicated by the standard deviation. The standard deviation for the coordinates in 2D images and in the 3D coordinate system is shown on the graphs (see \autoref{fig:grafovi}) as the Euclidean value. The coordinates \textit{i} and \textit{j} represent positions on a 2D image in pixel values, where \textit{i} denotes the horizontal coordinate(x) and \textit{j} denotes the vertical coordinate(y). Similarly, in a 3D space (in this case, the camera coordinate system), \textit{x} represents the vertical coordinate, \textit{y} represents the horizontal component and \textit{z} represents the distance from the landmark.

\noindent  To demonstrate the correlation between standard deviation and angle, the Spearman rank correlation coefficient (\autoref{eq:3}) was calculated for each landmark in the corresponding direction. In the given \autoref{eq:3}, \(\rho\) represents the Spearman rank correlation coefficient, \(d_i\) is the difference between the ranks of the two variables (here: standard deviation and angles) for each data point, and \textit{n} is the number of data points.
  \begin{equation}
      \rho = 1 - \frac{6 \sum d_i^2}{n(n^2 - 1)}
      \label{eq:3}
  \end{equation}
   The results of the Spearman rank correlation coefficient are as follows: for landmark T1, it is 0.9; for landmark T2, it is 0.7; for landmark T5, it is 0.2; and for landmark T7, it is 0.8. Coefficient values greater than 0.9 indicate a very strong positive correlation between the variables (standard deviations and angles). Coefficients 0.7 and 0.8 show a strong positive correlation, while a coefficient of 0.2 indicates a weak positive correlation.
   
\noindent Tracking the position and coordinate values of facial landmarks on a 2D image and their 3D coordinates, it is observed that with surgical light, higher rotation angles were achieved at which detection was possible. Surgical light provides better highlighting of the anatomical features of the face compared to front light on the camera. At maximal rotation angles (70° in yaw and 40° in pitch), the landmarks on the rotated side of the face remain more accurately localized, while those on the opposite side became increasingly scattered, which is further confirmed by the fact that the Spearman rank correlation coefficient for landmarks T1, T2 and T7 is greater than 0.7 (strong positive correlation).

\section{Conclusion}
In this paper, we tested the MediaPipe algorithm for facial landmark detection at various camera rotation angles under constant surgical lighting. The research demonstrated that lighting plays an significant role in maximizing the rotation angle for facial landmark detection, with surgical lighting enabling greater rotation angles compared to its absence. It is crucial to accurately define the direction and intensity of the light to best highlight the geometry of facial anatomical features. Moreover, the coordinates of specific facial landmarks vary with changes in the rotation angle and lighting conditions, with an increase in the rotation angle resulting in greater dispersion, while some landmarks become scattered across the face. Due to this scattering, the T5 landmark could not be consistently extracted from the disparity map at every position, resulting in missing 3D coordinate data. This is why the correlation for this landmark is 0.2, whereas for the other landmarks located on the side of the face towards which the camera rotated, the correlation remains strong. To assume a normal distribution, more data are required for each point (T1, T2, T5, T7) in yaw and pitch. In the upcoming research on this topic, a larger dataset is planned to enable more detailed statistical analysis, including p-values and confidence intervals.
Future research will focus on the development of a method for locating the patient's face with precision in the operating room. The system should be able to detect objects such as the operating table and lighting within the operating room under different lighting conditions, enabling it to recognize the patient. Regardless of the patient's position, the system should localize their face during the procedure and automatically adjust the position of the equipment (robot, camera) to ensure face detection and visibility of the surgical instruments. In this way, the processes within the operating room could be automated, reducing the risk to the patient during the procedure.

\section{Acknowledgement}
This research was funded by the project INSPIRATION – non-INvaSive PatIent RegistrATIon for rObotic Neurosurgery (grant no. NPOO.C3.2.R2-I1.06.0153), financed by the European Union through the National Recovery and Resilience Plan (NPOO).

\end{document}